\documentclass[conference]{IEEEtran}
\IEEEoverridecommandlockouts
\usepackage{cite}
\usepackage[utf8]{inputenc} 
\usepackage[T1]{fontenc}    
\usepackage{hyperref}       
\usepackage{url}            
\usepackage{booktabs}       
\usepackage{amsfonts}       
\usepackage{nicefrac}       
\usepackage{microtype}      
\usepackage{amsmath}
\usepackage{amssymb,bm,graphicx}
\usepackage[ruled,vlined]{algorithm2e}

\def\BibTeX{{\rm B\kern-.05em{\sc i\kern-.025em b}\kern-.08em
    T\kern-.1667em\lower.7ex\hbox{E}\kern-.125emX}}

\DeclareMathAlphabet\ten{OMS}{cmsy}{b}{n}
\DeclareMathAlphabet\mathcalbf{OMS}{cmsy}{b}{n}

\DeclareMathAlphabet\mathcalbf{OMS}{cmsy}{b}{n}

\begin{document}

\title{Graph Theory for Metro Traffic Modelling}


\author{\IEEEauthorblockN{Bruno Scalzo Dees, Yao Lei Xu, Anthony G. Constantinides, Danilo P. Mandic}
\IEEEauthorblockA{\textit{Department of Electrical and Electronic Engineering} \\
\textit{Imperial College London}\\
London, United Kingdom \\
\{bruno.scalzo-dees12, yao.xu15, a.constantinides, d.mandic\}@imperial.ac.uk}
}

\maketitle

\begin{abstract}
A unifying graph theoretic framework for the modelling of metro transportation networks is proposed. This is achieved by first introducing a basic graph framework for the modelling of the London underground system from a diffusion law point of view. This forms a basis for the analysis of both station importance and their vulnerability, whereby the concept of graph vertex centrality plays a key role. We next explore k-edge augmentation of a graph topology, and illustrate its usefulness both for improving the network robustness and as a planning tool. Upon establishing the graph theoretic attributes of the underlying graph topology, we proceed to introduce models for processing data on such a metro graph. Commuter movement is shown to obey the Fick's law of diffusion, where the graph Laplacian provides an analytical model for the diffusion process of commuter population dynamics. Finally, we also explore the application of modern deep learning models, such as graph neural networks and hyper-graph neural networks, as general purpose models for the modelling and forecasting of underground data, especially in the context of the morning and evening rush hours. Comprehensive simulations including the passenger in- and out-flows during the morning rush hour in London demonstrates the advantages of the graph models in metro planning and traffic management, a formal mathematical approach with wide economic implications.
\end{abstract}


\section{Introduction}

The rapid development of many world's economies has been followed by an increasing proportion of population moving to cities \cite{kuz2016graph, derrible2009network}. Through a "domino-effect" of such a huge societal change, world's urban traffic congestion has become, and is likely to remain, an overwhelming issue. In addition, underground traffic networks frequently experience signal failures and train derailments - not to mention emergency measures due to accidents - while closure of one station may impact the service across the entire network. The economic costs of such transport delays to central London business are estimated to be $\pounds 1.2$ billion per year. Appropriate and physically meaningful tools to understand, quantify, and plan for the resilience of transportation networks to disruptions are therefore a pre-requisite for the planning and daily running of public transport, a subject of this work.

Graph representations for traffic planning have a long history \cite{badger2012evolution, gattuso2005compared, stoilova2015application, guze2014graph, boulmakoul2017combinatorial}; as far back as in 1926, a map-maker named Fred Stingemore produced a map of the London underground by regularising the spacing between stations, while allowing himself some artistic freedom with the routes of the various lines. Stingemore's work was further abstracted by Harry Beck in 1933 to the graph form we have today. In this article, we move beyond the graphs themselves and employ the concepts of Data on Graph \cite{stankovic2019graph, stankovic2019graphII, stankovic2020graphIII}, to address some burning issues in underground public transport, such as "sensitivity" of the network to station closures. We further demonstrate how graph theory can be used to identify those stations in the London underground network which have the greatest influence on the functionality of the traffic, and proceed, in an innovative way, to assess the impact of a station closure on service levels across the city. Such underground network vulnerability analysis is shown to provide the opportunity to analyse, optimize and enhance the connectivity of the traffic networks in a mathematically tractable and physically meaningful way. We also explore the benefits of modelling the London underground graph via modern deep learning approaches, through leveraging models such as graph neural network and hyper-graph neural networks to perform semi-supervised learning on graphs.



\section{Traffic Centrality and Vulnerability}

\begin{figure}[h!]
	\centering
	\vspace{-8mm}
	\includegraphics[width=1\columnwidth]{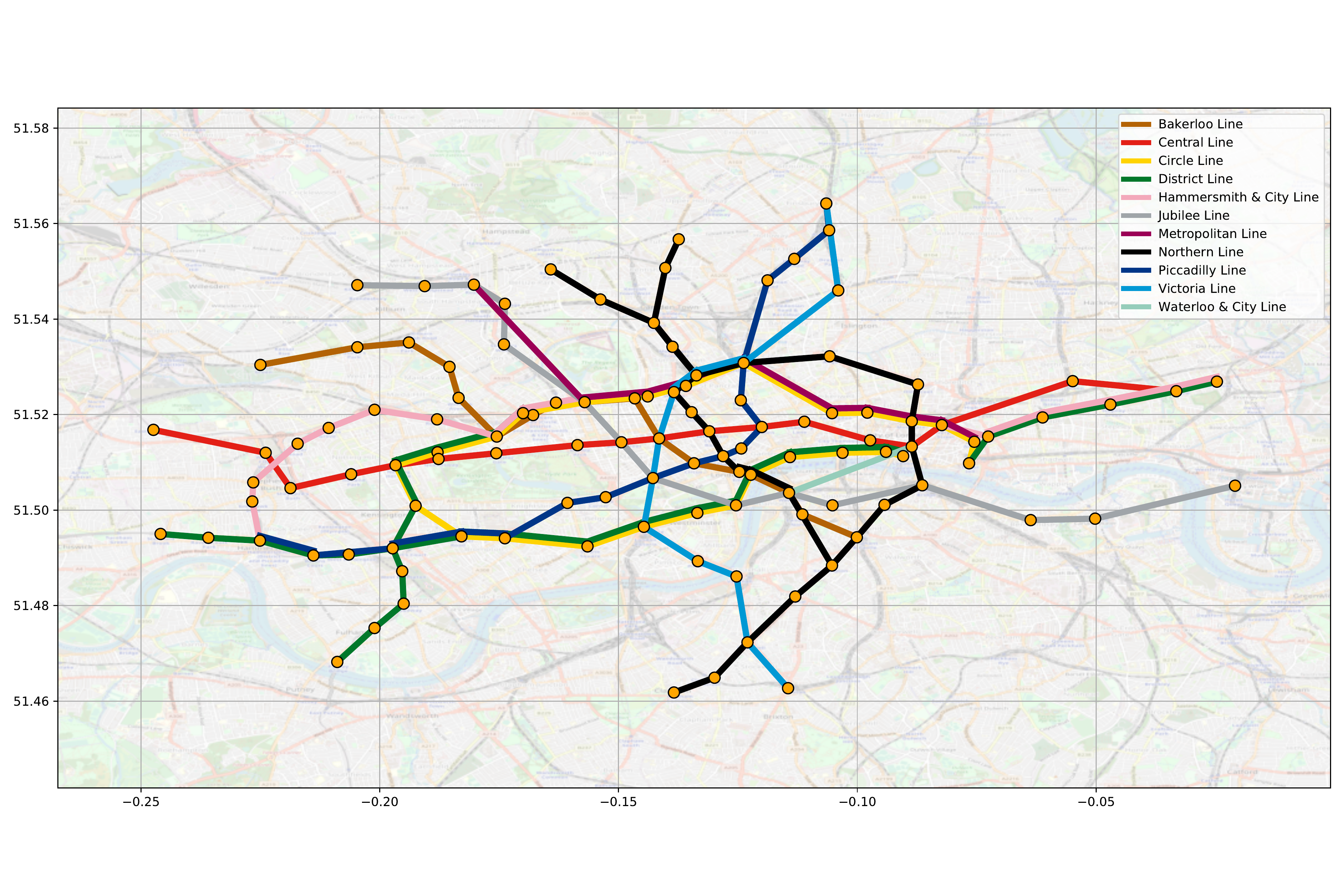}
	\vspace{-12mm}
	\caption{\label{fig:underground_graph}{Graph representation of the London underground network in Zones 1 and 2. The circles denote the vertices (stations) and the lines between the circles designate the underground lines.}}
\end{figure}

We consider the underground network as an undirected $N$-vertex graph, denoted by $\mathcal{G} = \{ \mathcal{V}, \mathcal{E} \}$, with $\mathcal{V}$ as the set of $N$ vertices (stations) and $\mathcal{E}$ the set of edges (underground lines) connecting the vertices (stations). The connectivity of the network is designated by the (undirected) adjacency matrix, $\mathbf{A} \in \mathbb{R}^{N \times N}$, where $a_{ij} = 1$ if a connection between the $i$-th and $j$-th station exists, and 0 otherwise  \cite{stankovic2019graph}. Figure \ref{fig:underground_graph} shows the proposed graph model of the London underground network, with each vertex representing a station, and each edge designating the the connection between two adjacent stations through one or more underground lines. For visualization purposes, we show the graph representation for the central zones 1-2, out of the 6 zones of the London underground network. Notice that standard Data Analytics typically assume regular grids in time and in space, and are thus ill-equipped to deal with this class of problems, which reside on irregular domains that may exhibit multiple paths between adjacent vertices (stations). This inadequacy becomes even more prominent during the "anomaly" times, such as the morning an evening rush hours.

\begin{figure}[h!]
	\centering
	\vspace{-8mm}
	\includegraphics[width=0.99\columnwidth]{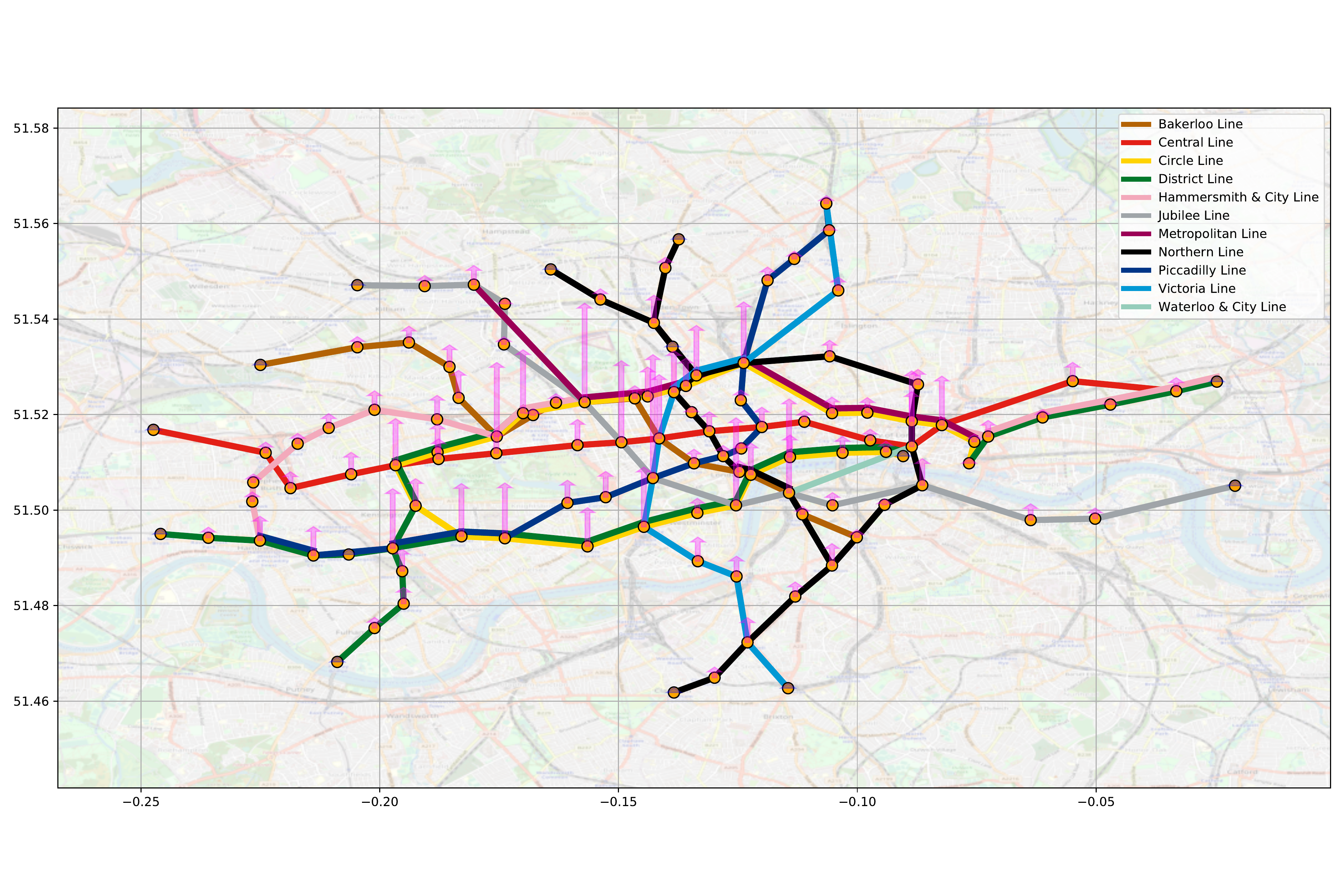}
	\vspace{-12mm}
	\caption{\label{fig:betweenness}{Betweenness centrality, designated by magenta-coloured bars, of the London underground network in Zones 1 and 2.}}
\end{figure}

The graph representation of the traffic network allows us to explore the usefulness of the \textit{betweenness centrality} metric as a measure of the vulnerability of a certain station to traffic disruption. This metric is quite natural in the context of transportation planning, as the betweenness centrality measures the extent to which a given vertex lies in between pairs or groups of other vertices of the graph, and is given by
\begin{align}
	B_{n} = \sum_{k,m \in \mathcal{V}} \frac{\sigma(k,m|n)}{\sigma(k,m)}
\end{align}
where $\sigma(k,m)$ denotes the number of shortest paths between vertices $k$ and $m$, and $\sigma(k,m|n)$ the number of those paths passing through a vertex $n$ \cite{Freeman1977}. In other words, betweenness centrality reflects the extent to which a vertex (station) is an intermediate in the communication over the graph (underground network). Figure \ref{fig:betweenness} depicts the values of betweenness centrality of stations in the London underground network in Zones 1 and 2. Observe that, as expected, the stations at the centre of the city exhibit the largest betweenness centrality, while the stations in the outer residential area have the smallest betweenness centrality. Physically, this means that disruptions to lines passing through stations with high betweenness centrality would cause less impact on the overall network, as there are alternative ways of accessing the same station via other underground lines. Conversely, if connections to low centrality stations were to be disrupted, it would cut-off both that station and those further out from accessing the rest of the network.

	
\begin{table}[h!]
	\setlength{\tabcolsep}{7pt}
	\renewcommand{\arraystretch}{1.2}
	\begin{center}
		\caption{\label{table:centrality_top_3}{Stations with largest betweenness centrality.}}
		\begin{tabular}{l|r r r}
			\hline
			\textbf{Station} & \textbf{Entries} & \textbf{Exits} & \textbf{Net out-flow} \\
			\hline
			Bank          & $17,577$   & $69,972$  & $52,395$   \\
			Canary Wharf  & $8,850$    & $56,256$  & $47,406$   \\
			Oxford Circus & $3,005$    & $44,891$  & $41,886$   \\ \hline
		\end{tabular}
		\vspace{-6mm}
	\end{center}
\end{table}

\section{K-Edge Augmentation for Robust Transport Networks}

Given that the partial closure of a line between any two metro stations will severely impact the entire metro system \cite{anop2016using}, we set out to analyse the vulnerability of the metro system in terms of the level of connectivity, and proceed to explore possible improvements towards increased resilience.

To this end, we introduce a measure of the underground network robustness in terms of connectivity; this is achieved through the concept of \textit{$k$-edge connectivity}. Formally, a graph is said to be $k$-edge connected if it cannot be regrouped into distinct sub-graphs unless $k$ or more edges are removed. For the London underground network in Zones 1 and 2 in Figure \ref{fig:underground_graph}, the graph topology forms a $1$-edge connected graph. This implies that a removal of even one connection (between two adjacent stations) suffices to disconnect some stations from the rest of the network. It then comes as no surprise that the stations exhibiting lowest betweeness centrality (see Figure \ref{fig:betweenness}) are most vulnerable to the closure of other stations.

\begin{figure}[h!]
	\centering
	\vspace{-8mm}
	\includegraphics[width=0.99\columnwidth]{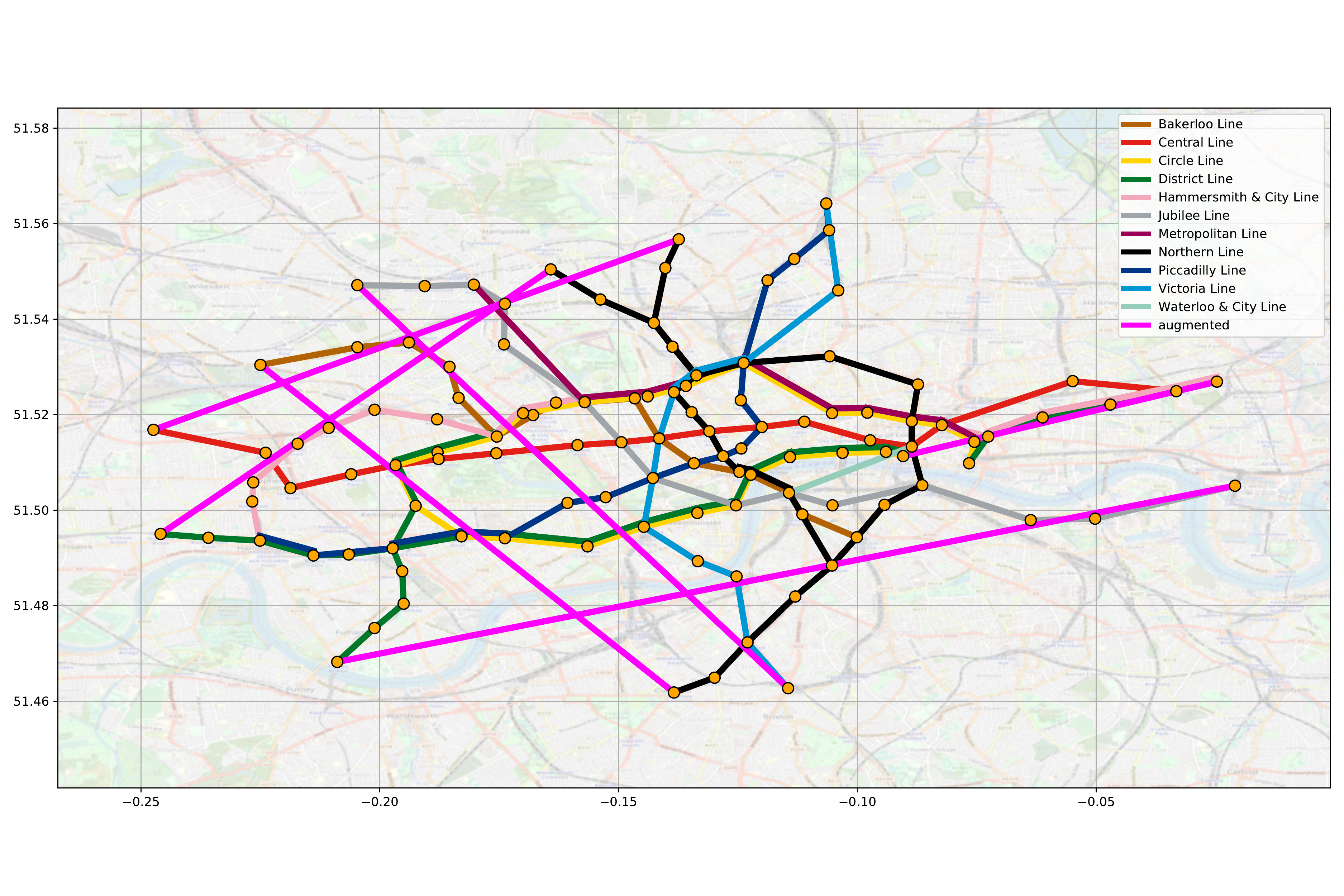}
	\vspace{-12mm}
	\caption{\label{fig:kedge_raw}{Graph of the London underground network in Zones 1 and 2, after performing a naive k-edge augmentation, for $k$=2.}}
\end{figure}

Our solution for improving the robustness of the overall network is based on \textit{$k$-edge augmentation}, a search problem for determining the minimum set of additional edges, $\mathcal{A}$, such that $\mathcal{A} \cap \mathcal{E} = \varnothing$ and the resulting graph $\mathcal{G}_k = (\mathcal{V}, \mathcal{E} \cup \mathcal{A})$ remains $k$-edge connected \cite{watanabe1987edge}. Figure \ref{fig:kedge_raw} shows an example of $k$-edge augmentation for the London underground network for $k=2$. 

\begin{figure}[h!]
	\centering
	\vspace{-8mm}
	\includegraphics[width=0.99\columnwidth]{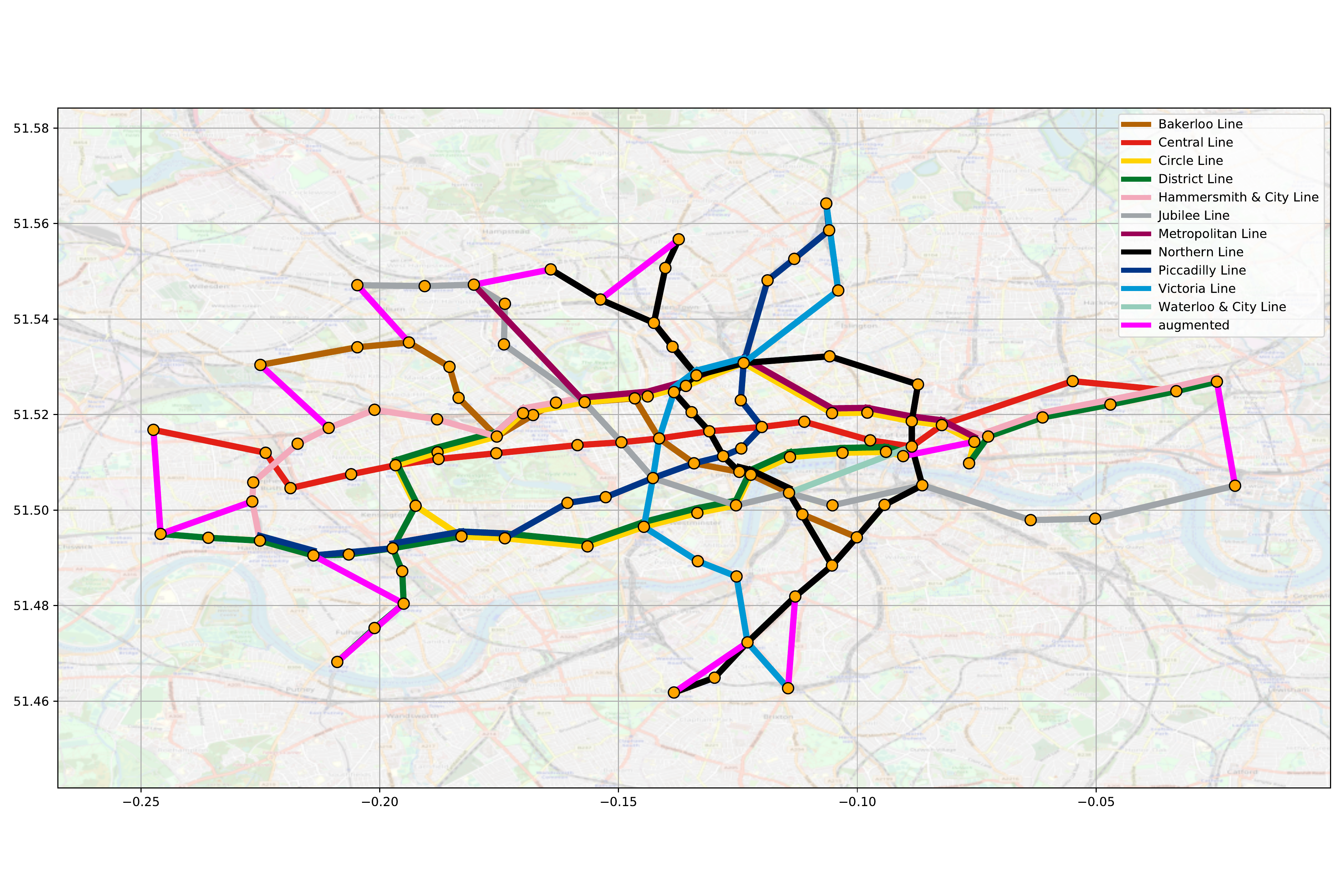}
	\vspace{-12mm}
	\caption{\label{fig:kedge_constrained}{Graph of the London underground network in Zones 1 and 2 after performing geographically constrained k-edge augmentation, for $k$=2.}}
\end{figure}

Although the solution presented in Figure \ref{fig:kedge_raw} is optimal in terms of the number of new connections required, $|\mathcal{A}|$, it is highly unrealistic to build such direct long-range connections between stations. To this end, assuming that the cost of building a new connection is proportional to the geographic distance between two stations, we can constrain the search space of the k-edge augmentation problem so as to include only new connections which are quite short, that is $d(v_1, v_2) < \alpha$ for all $(v_1, v_2) \in \mathcal{A}$, where $d(\cdot)$ denotes the geographic distance between any two stations and $\alpha$ is a user chosen threshold. Figure \ref{fig:kedge_constrained} shows such a physically more realistic augmented network, with significantly shorter new connections.

\section{Modelling Commuter Population from Net Passenger Flow}

\begin{figure}[h!]
	\centering
	\vspace{-2mm}
	\includegraphics[width=0.3\textwidth, trim={0 3cm 0 4cm}, clip]{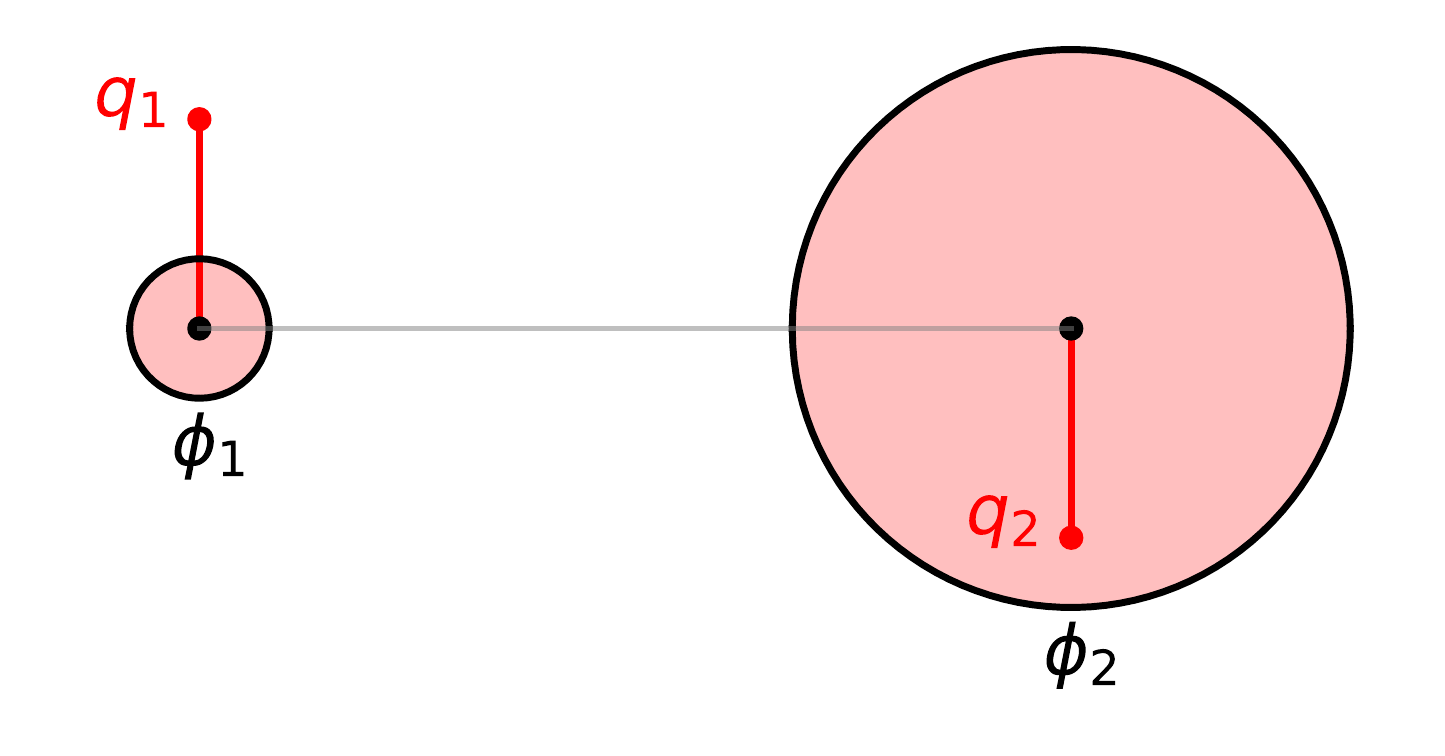}
	\vspace{-2mm}
	\caption{\label{fig:ficks_law}{A graph representation of the London underground network through Fick's law of diffusion. Consider a simplified path graph with two stations surrounded by the respective populations, $\phi_{1}$ and $\phi_{2}$ (proportional to the circle area), which exhibit the corresponding net fluxes, $q_{1}$ and $q_{2}$. Stations surrounded by large population (residential areas) exhibit a net in-flow of passengers, while stations surrounded by low population (business districts) experience net out-flow of passengers. The overall net flow (in-flow and out-flow) of passengers across the entire network sums up to zero.}}
\end{figure}

We shall now demonstrate that the graph theoretic approach to transportation modelling allows us to answer even broader societal questions, such as to infer the resident population surrounding each station solely based on the net passenger flow during the morning rush hour.

To derive the corresponding graph model, we employ the \textit{Fick's law of diffusion} which relates the diffusive flux to the concentration of a given vector field, under the assumption of a steady state. According to Fick's law, the flux flows from regions of high concentration (population) to regions of low concentration (population), with a magnitude proportional to the concentration gradient. Mathematically, the Fick law is given by
\begin{align} \label{eq:fickslaw_eq}
	\mathbf{q} = - k \nabla \boldsymbol{\phi}
\end{align}
where 
$\mathbf{q}$ is the flux which measures the amount of substance per unit area per unit time (mol m$^{-2}$ s$^{-1}$), $k$ is the coefficient of diffusivity measured in area per unit time (m$^{2}$ s$^{-1}$), while $\boldsymbol{\phi}$ represents the concentration (mol m$^{-3}$).

This framework allows us to model the passenger flow in the London underground network as a diffusion process, whereby during the morning rush hour the population mainly flows from population-dense residential areas to sparsely populated business districts. Based on the general Fick's law in (\ref{eq:fickslaw_eq}), the variables in our proposed metro traffic model are: (i) $\mathbf{q} \in \mathbb{R}^{N}$, the net passenger flow vector, with the $i$-th entry as the net passenger flow at the $i$-th station during the morning rush hour, given by: $q_{i} = \ln{({p_i^{(o)}})} - \ln{({p_i^{(i)}})}$, where $p_i^{(o)}$ and $p_i^{(i)}$ denote respectively the number of passengers exiting and entering the $i$-th station, with its dimension equal to ``passengers per station per unit time''; (ii) $k=1$, the coefficient of diffusivity, with its dimension equal to ``stations per unit time''; (iii) $\boldsymbol{\phi} \in \mathbb{R}^{N}$, which represents the resident population in the area surrounding the station. 

The proposed graph-theoretic model is physically meaningful and suggests that, in the morning, the net passenger flow at the $i$-th station, $q_{i}$, is proportional to the population difference between the areas surrounding a station $i$ and the adjacent stations $j$, that is
\begin{equation}
    \begin{split}
    	q_{i} 
    	&= - k \sum_{j} A_{ij}(\phi_{i} - \phi_{j}) \\
    	&= - k \left( \phi_{i} \sum_{j} A_{ij} - \sum_{j} A_{ij}\phi_{j} \right)  \\
    	&= - k \left( \phi_{i} D_{ii} - \sum_{j} A_{ij} \phi_{j} \right) \\
    	&= - k \sum_{j}\left( \delta_{ij} D_{ii} - A_{ij} \right) \phi_{j} \\
    	&= - k \sum_{j} L_{ij} \phi_{j}
    \end{split}
\end{equation}

The model in (\ref{eq:fickslaw_eq}) also allows us to jointly consider $N$ stations, to yield a matrix form
\begin{align}
	\mathbf{q} = -k\mathbf{L}\boldsymbol{\phi} \label{eq:graph_Fick_law}
\end{align}
where $\mathbf{L} = (\mathbf{D}-\mathbf{A}) \in \mathbb{R}^{N \times N}$ is the Laplacian matrix of the graph in Figure \ref{fig:underground_graph} \cite{stankovic2019graphII}. Also, the commuter dynamics model based on Fick's law has an intimate connection with the graph shift operators \cite{stankovic2019graph, stankovic2019graphII}. For enhanced physical intuition, Figure \ref{fig:ficks_law} shows a signal within this diffusion model on a $2$-vertex path graph obeying the Fick law. 

\begin{figure}[h!]
	\centering
	\vspace{-8mm}
	\includegraphics[width=0.99\columnwidth]{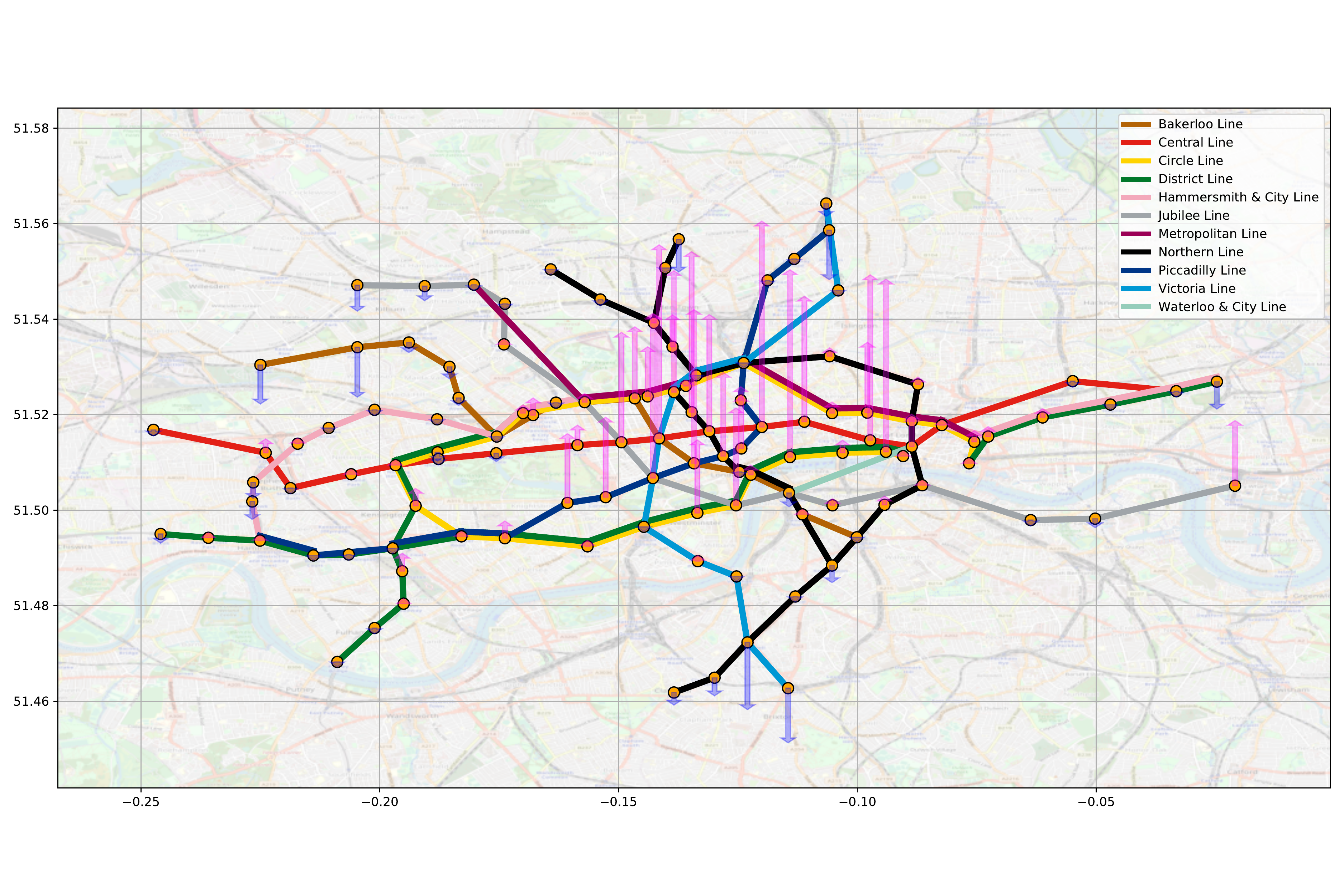}
	\vspace{-12mm}
	\caption{\label{fig:tube_net_flow}{Net passenger out-flow (in magenta) and in-flow (in blue) during the morning rush hour.}}
\end{figure}

The data for the average daily net flow of passengers during the morning rush hour at each station in $2016$ was obtained from Transport for London (TFL) \cite{TFL}, and is visualised as a signal on the London underground graph model in Figure \ref{fig:tube_net_flow}. 


Table \ref{table:netflows_top5} shows the average net flow of passengers for the top $5$ stations with the greatest net in-flow and out-flow. Observe that the stations which exhibits largest net out-flow of passengers are located within the financial (Bank, Canary Wharf, Green Park) and commercial (Oxford Circus, Holborn) districts. In contrast, the stations exhibiting the greatest net in-flow of passengers are located in residential areas.

\begin{table}[h!]
	\setlength{\tabcolsep}{7pt}
	\renewcommand{\arraystretch}{1.2}
	\begin{center}
		\caption{\label{table:netflows_top5}{Stations with largest net passenger out-flow and in-flow.}}
		\begin{tabular}{l|r r r}
			\hline
			\textbf{Station} & \textbf{Entries} & \textbf{Exits} & \textbf{Net out-flow} \\
			\hline
			Bank          & $17,577$   & $69,972$  & $52,395$   \\
			Canary Wharf  & $8,850$    & $56,256$  & $47,406$   \\
			Oxford Circus & $3,005$    & $44,891$  & $41,886$   \\ \hline
			Finsbury Park & $20,773$   & $8,070$   & $- 12,703$ \\
			Canada Water  & $31,815$   & $14,862$  & $- 16,953$ \\
			Brixton       & $24,750$   & $4,369$   & $- 20,381$ \\ \hline
		\end{tabular}
		\vspace{-2mm}
	\end{center}
\end{table}

\begin{figure}[h!]
	\centering
	\vspace{-8mm}
	\includegraphics[width=1\columnwidth]{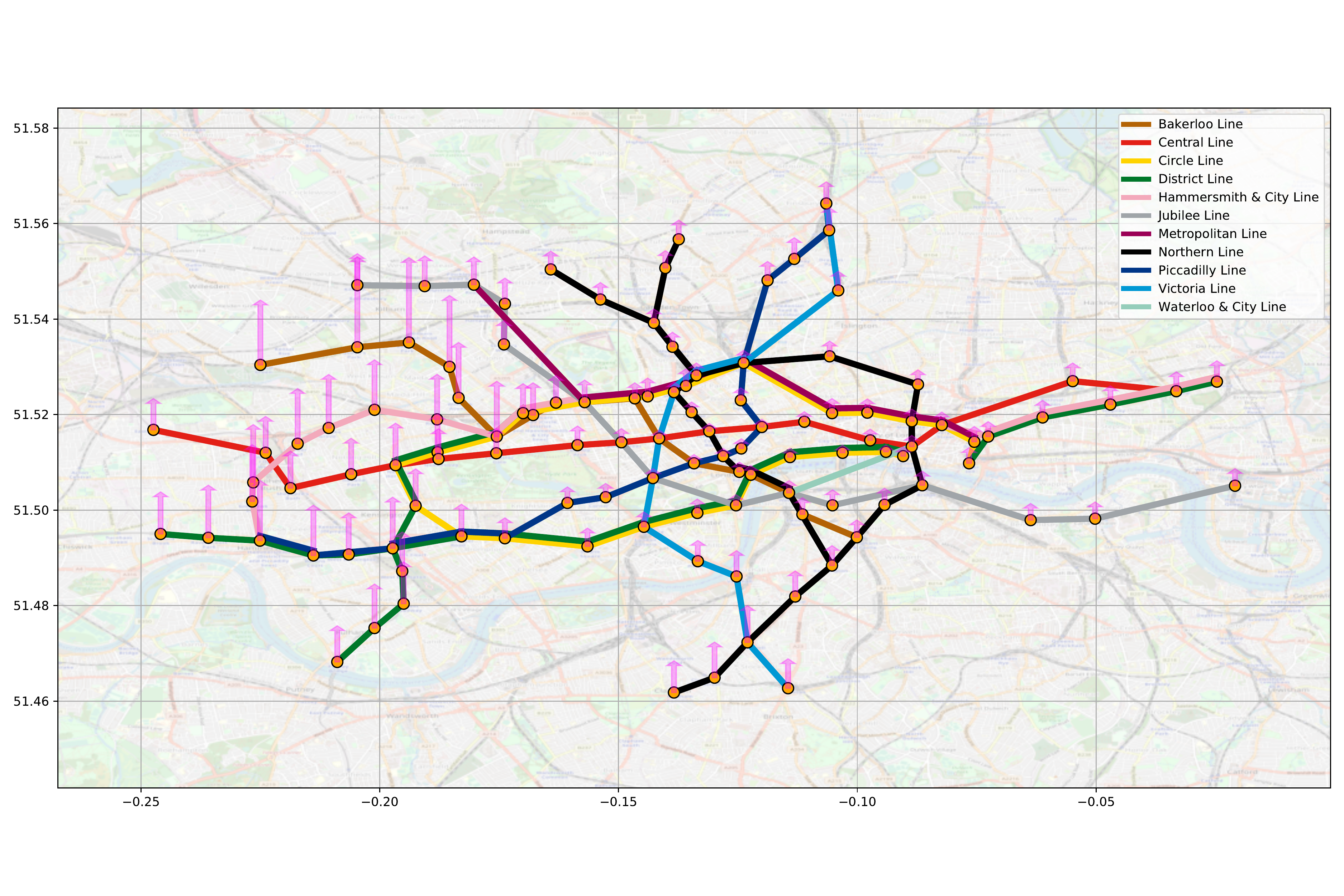}
	\vspace{-12mm}
	\caption{\label{fig:population}{Population density implied by our graph model in (\ref{eq:population_estimate}), obtained from the net passenger out-flow during the morning rush hour within Zones 1 and 2. As expected, business districts exhibit the lowest population density, while residential areas (Zone 2) exhibit the highest commuter population density.}}
\end{figure}

To obtain an estimate of the resident population surrounding each station, recall that the vector of populations, $\boldsymbol{\phi} \in \mathbb{R}^{N}$, is subject to (\ref{eq:graph_Fick_law}), that is
\begin{align}
	\hat{\boldsymbol{\phi}} = -\frac{1}{k} \mathbf{L}^{+}\mathbf{q} \label{eq:population_estimate}
\end{align}
where the symbol $(\cdot)^{+}$ denotes the matrix pseudo-inverse operator. It is important to mention that the population vector can only be estimated up to a constant, hence the vector $\hat{\boldsymbol{\phi}}$ actually quantifies the \textit{relative} population between stations, whereby the station with the lowest estimated surrounding population takes the value of $0$. The so estimated resident population, based on the morning net passenger flow, is displayed as a signal on a graph in Figure \ref{fig:population}. Observe that the so obtained estimates are reasonable, since most of the resident population in London is concentrated toward the more remote areas of Zone 2, while business districts at the centre of Zone 1 are sparsely populated in the evening.

\section{A Hypergraph Model of Metro Network}

The analysis of the London metro system carried out so far has been based on the classical graph adjacency matrix, $\textbf{A} \in \mathbb{R} ^ {N \times N}$ (and the corresponding graph Laplacian matrix $\textbf{L} = \textbf{D} - \textbf{A}$), where $A_{ij} = 1$ if a connection exists between the $i$-th and $j$-th station, and $0$ otherwise. However, despite its ability to capture the underlying network topology, the classical graph adjacency matrix cannot account for higher order connectivity information, such as multiple connections between the adjacent stations. To this end, we here explore the hypergraph representation of the underground network, which naturally accounts for higher order connectivity in a graph.

A hypergraph \cite{feng2019hypergraph}, $\mathcal{H} = (\mathcal{V}, \mathcal{E})$, is a generalization of the classical graph, and is characterized by a vertex set, $\mathcal{V}$, and a hyperedge set, $\mathcal{E}$, whereby each hyperedge can contain any number of vertices. To compactly represent such a hypergraph, we shall represent $\mathcal{H}$ with the "vertex-edge" incidence matrix, $\mathbf{M} \in \mathbb{R}^{|\mathcal{V}| \times |\mathcal{E}|}$, such that $M_{v, e} = 1$ if $v \in e$ and $0$ otherwise. Notice the difference from the "vertex-vertex" adjacency matrix, \textbf{A}. The degree of a vertex in a hypergraph is defined as $d(v) = \sum_{e \in \mathcal{E}} M_{v,e}$, while the degree of an edge is defined as $d(e) = \sum_{v \in \mathcal{V}} M_{v,e}$. 

\begin{figure}[h!]
	\centering
	\includegraphics[width=0.8\columnwidth]{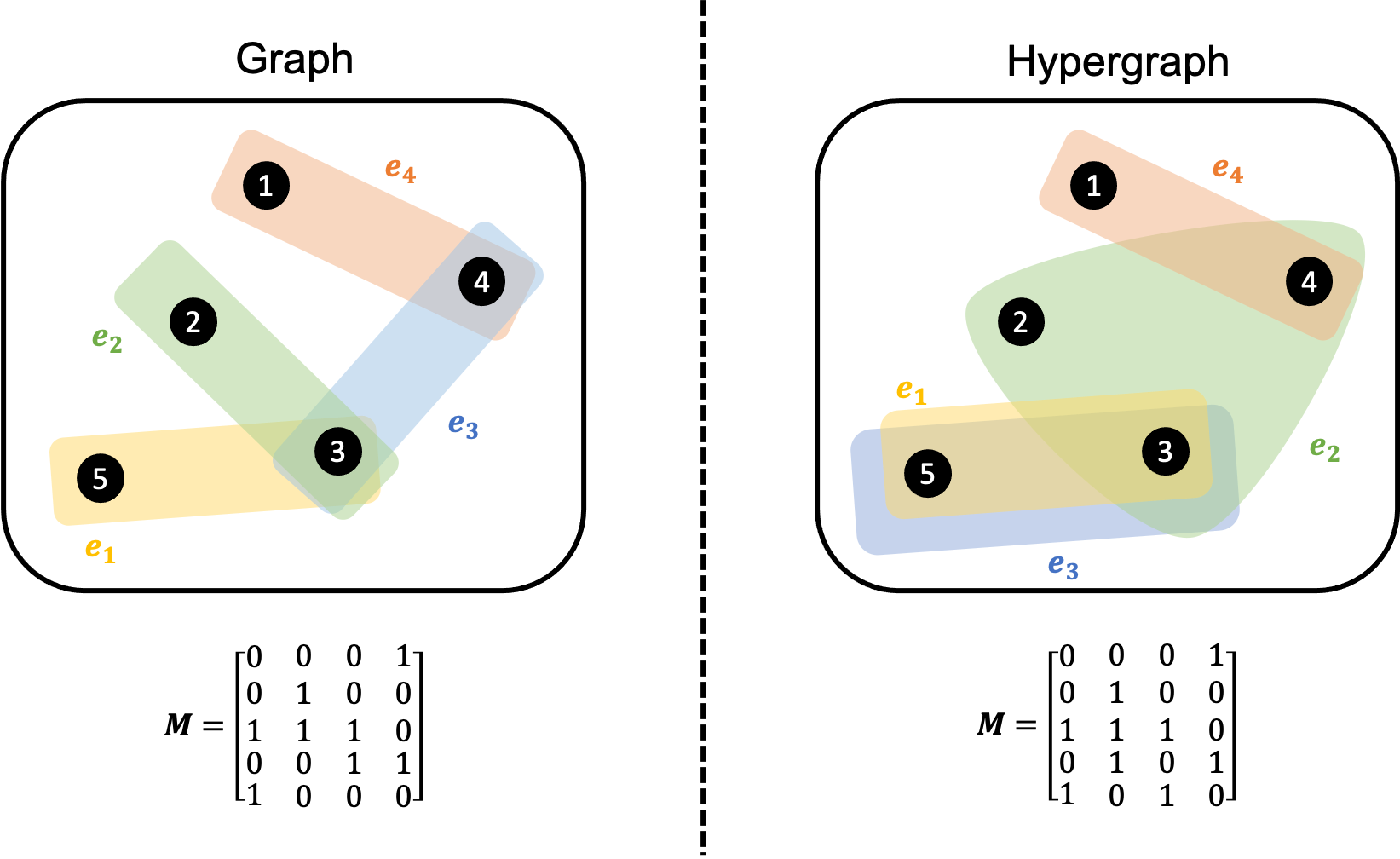}
	\caption{\label{fig:hypergraph}{Graphs (left) vs Hypergraphs (right)}}
\end{figure}

The hypergraph matrix of a normalized undirected hypergraph can then be defined as
\begin{equation}
    \boldsymbol{\Theta} = \textbf{D}_v^{-\frac{1}{2}} \textbf{M} \textbf{D}_e^{-1} \textbf{M}^T \textbf{D}_v^{-\frac{1}{2}}
\end{equation}
where $\textbf{D}_v \in \mathbb{R}^{|\mathcal{V}| \times |\mathcal{V}|}$ is the \textit{vertex degree} diagonal matrix composed of the individual vertex degrees, $d(v)$, for all vertices $v \in \mathcal{V}$, and $\textbf{D}_e \in \mathbb{R}^{|\mathcal{E}| \times |\mathcal{E}|}$ is the \textit{edge degree} diagonal matrix composed of edge degrees, $d(e)$, for all $e \in \mathcal{E}$. The normalised hypergraph Laplacian matrix is defined as $\boldsymbol{\Delta} = \textbf{I} - \boldsymbol{\Theta}$ \cite{feng2019hypergraph}.

To construct the hypergraph representation for the metro network, we shall simply append all multi-modal connections of the system (i.e. connections across different metro lines) as additional edges in the incidence matrix.


To validate the expressive power of the hypergraph modelling, we employ a semi-supervised regression experiment, where we estimate the net passenger out-flow during the evening rush hour for each of the underground stations, given the morning passenger flow data and the network connectivity structure. More specifically, given an input matrix, $\textbf{X} \in \mathbb{R}^{N \times F}$, with $F$ as passenger flow features for each of the $N$ stations, we build a neural network model that estimates the vector, $\textbf{y} \in \mathbb{R}^N$, of which each entry, $y_i$, corresponds to the evening rush hour net passenger out-flow for the $i$-th station. 

The proposed neural network model implements a first order spatial hypergraph filter given by
\begin{equation}
    \textbf{Y} = \sigma \big( \left(\textbf{I} + \alpha \boldsymbol{\Theta} \right) \textbf{X} \textbf{W} \big)
\end{equation}
where $\alpha$ is a learnable graph filter coefficient, $\textbf{W}$ is a learnable weight matrix performing feature extraction, and $\sigma$ is an activation function. In a classical graph neural network implementation, we simply replace the hypergraph matrix, $\boldsymbol{\Theta}$, with the standard adjacency matrix, $\textbf{A}$ \cite{stankovic2020graphIII}.

\begin{figure}[h!]
	\centering
	\vspace{-4mm}
	\includegraphics[width=0.99\columnwidth]{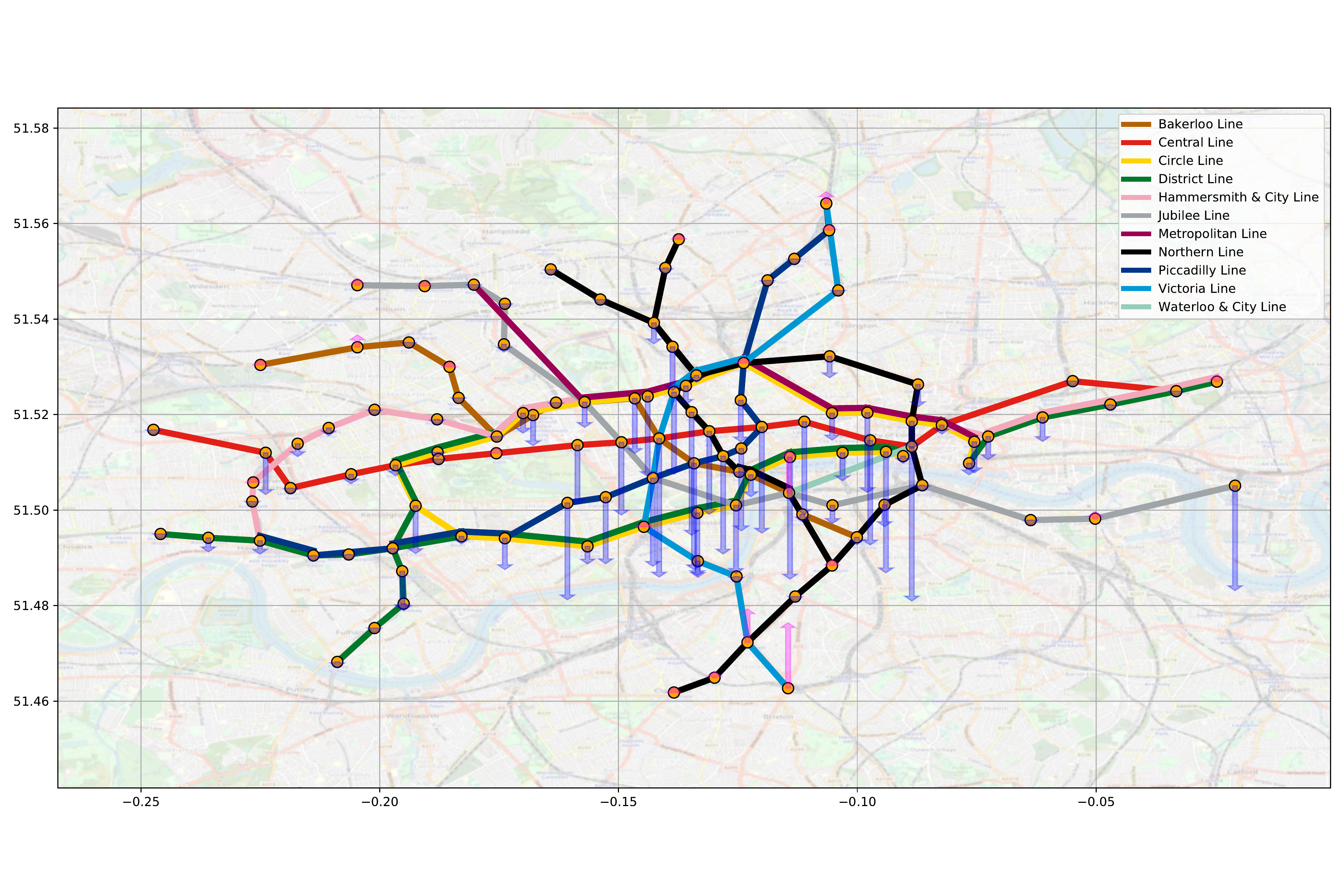}
	\vspace{-8mm}
	\caption{\label{fig:tube_transportation_modalities}{The evening net passenger-flow, estimated from the morning rush hour data, based on the hypergaph neural network model. The power of the proposed hypergraph model is reflected in the reverse net flow from the morning rush hours, as people move from central business areas back to residential zones.}}
\end{figure}

Table \ref{table:experiment_res} shows the corresponding Mean Squared Error (MSE), with both graph neural network models out-performing the standard neural network in the prediction of the evening traffic based on the morning rush hour data. The Hypergraph graph neural network model exhibited the best performance, verifying its expressivity and ability to compactly capture the multi-modal connectivity information. Figure \ref{fig:tube_transportation_modalities} shows the prediction results from the proposed experiment using a hypergraph model, where the estimated evening passenger flow for each of the station is plotted on the underground graph. When measuring the out-of-sample performance, the hypergraph model achieved the lowest MSE, shown in Table \ref{table:experiment_res}, hence demonstrating its ability to capture multi-way connections between stations.

\begin{table}[h!]
	\setlength{\tabcolsep}{7pt}
	\renewcommand{\arraystretch}{1.2}
	\begin{center}
		\caption{\label{table:experiment_res}{Semi-supervised learning results from the proposed experiment.}}
		\begin{tabular}{c|rrr}
			\hline
			\textbf{Model} & \textbf{Training MSE} & \textbf{Testing MSE} \\
			\hline
			Neural Network    & $0.2736$ & $0.3171$ \\
			Graph Neural Network    & $\textbf{0.2383}$ & $0.2750$ \\
			Hypergraph Neural Network    & $0.2400$ & $\textbf{0.2683}$ \\
			\hline
		\end{tabular}
		\vspace{-2mm}
	\end{center}
\end{table}

\section{Conclusion}
\label{sec:conclusion}

We have introduced a graph-theoretic framework which makes it possible to assess the functional design and operation of public transport networks in a mathematically tractable and intuitive manner. The formulation allows not only for an assessment of the functionality, but also offers an opportunity to facilitate enhancements and assess vulnerabilities, notwithstanding the possibility of a dynamic management of the network. Our focus has been on establishing a link between the laws of diffusive fields and the parameters of urban underground traffic. In this way, we have been able to leverage on the ability of graphs to represent data on irregular domains to introduce an intuitive model for the population migration during the rush hours. Furthermore, we have employed the hyper-graph representation of the network connections, and have demonstrated that this accounts for multiple links between underground stations in a natural way, which is not achievable based on the standard graph model. Finally, we have explored the connection between graphs and modern graph neural networks models in this context, demonstrating the power of this approach on the prediction of the evening traffic based on the morning traffic data. It is our hope to have provided a platform to approach both traffic planning and urban transition issues from a rigorous engineering perspective using graph signal processing techniques.

\bibliographystyle{IEEEtran.bst}
\bibliography{references.bib}

\end{document}